\documentclass[letterpaper, 10 pt, conference]{ieeeconf}  
\usepackage{graphicx}
\usepackage{amsmath,amsfonts}
\usepackage{booktabs}   
\usepackage{multirow}   
\usepackage{comment}  

\IEEEoverridecommandlockouts                              

\title{\LARGE \bf
LR-SGS: Robust LiDAR-Reflectance-Guided Salient Gaussian Splatting for Self-Driving Scene Reconstruction
}

\author{ 
Ziyu Chen, Fan Zhu, Hui Zhu, Deyi Kong, Xinkai Kuang, Yujia Zhang, Chunmao Jiang$^{*}$
\thanks{$^{*}$Corresponding author: Chunmao Jiang.}%
}

\begin{document}

\maketitle
\thispagestyle{empty}
\pagestyle{empty}

\begin{abstract}

Recent 3D Gaussian Splatting (3DGS) methods have demonstrated the feasibility of self-driving scene reconstruction and novel view synthesis. However, most existing methods either rely solely on cameras or use LiDAR only for Gaussian initialization or depth supervision, while the rich scene information contained in point clouds, such as reflectance, and the complementarity between LiDAR and RGB have not been fully exploited, leading to degradation in challenging self-driving scenes, such as those with high ego-motion and complex lighting. To address these issues, we propose a robust and efficient LiDAR-reflectance-guided Salient Gaussian Splatting method (LR-SGS) for self-driving scenes, which introduces a structure-aware Salient Gaussian representation, initialized from geometric and reflectance feature points extracted from LiDAR and refined through a salient transform and improved density control to capture edge and planar structures. Furthermore, we calibrate LiDAR intensity into reflectance and attach it to each Gaussian as a lighting-invariant material channel, jointly aligned with RGB to enforce boundary consistency. Extensive experiments on the Waymo Open Dataset demonstrate that LR-SGS achieves superior reconstruction performance with fewer Gaussians and shorter training time. In particular, on Complex Lighting scenes, our method surpasses OmniRe by 1.18 dB PSNR.
\end{abstract}

\section{INTRODUCTION}

High-fidelity reconstruction and novel view synthesis techniques for self-driving scenes are of significant value to the testing and training of end-to-end self-driving models. For model testing, they can reproduce the critical safety events that are difficult to handle in real-world driving scenes into the controllable digital space, enabling more flexible and safer retesting of improved algorithms. For model training, a single scenario enables diverse expansion to synthesize richer and more diverse training data, simulate data generation under new sensor arrangements, and obviate the requirement for re-collecting training data when transferring end-to-end self-driving models across different vehicles \cite{ref34}.

In recent years, reconstruction methods represented by image-based 3D Gaussian Splatting (3DGS)  \cite{ref1} have demonstrated fast and high-fidelity photorealistic rendering performance, and numerous studies have attempted to apply these methods to outdoor driving scenes, aiming to synthesize testing and training data for self-driving models under novel views. However, in self-driving scenes, camera-only methods \cite{ref8,ref11,ref15,ref37} are susceptible to complex lighting conditions and significant ego-motion, which leads to texture inconsistencies and unstable optimization. Notably, the multi-modal data captured by multiple onboard sensors are typically available. Fully exploiting this complementary information within a unified explicit representation to achieve accurate reconstruction and real-time rendering remains an open and critical challenge.

\begin{figure}[t]
	\centering
	\includegraphics[width=3.4in]{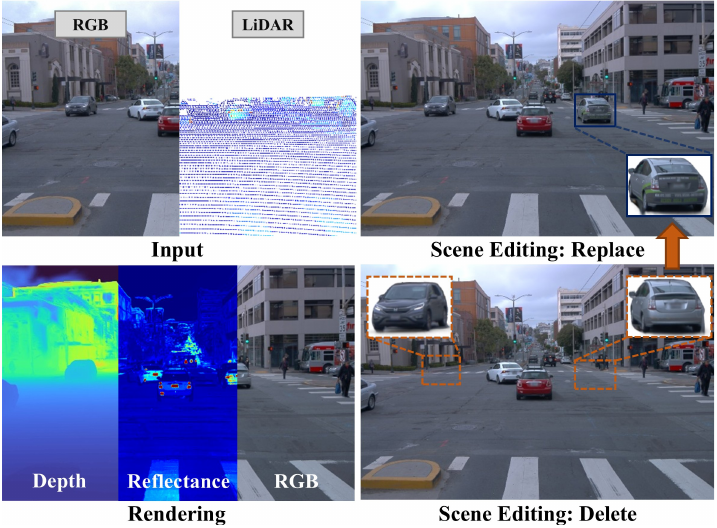}
	\caption{\textbf{Overview of LR-SGS.} Given RGB and LiDAR sequences as input, the method produces high-fidelity geometry, reflectance, and RGB renderings (left). Accurate modeling of background and objects enables realistic scene editing, including replacement and deletion (right).}
	\label{Overview_of_LRSGS}
\vspace{-2pt}
\end{figure}

In 3DGS-based scene reconstruction, relying solely on the RGB images captured by cameras cannot reliably ensure geometric and appearance consistency in reconstruction, as RGB signals are susceptible to disturbances from lighting, exposure, and other external factors. LiDAR provides accurate depth and is insensitive to lighting variations, making it a strong complement to RGB image information. Beyond depth, raw LiDAR returns also contain intensity measurements that can be calibrated into material-related reflectance, which is approximately lighting-invariant \cite{ref3, ref4}. However, most existing methods \cite{ref5, ref6, ref9, ref10} for integrating LiDAR into Gaussian Splatting rely on direct Gaussian initialization and depth supervision using point clouds, without fully exploiting the geometric and textural information carried by LiDAR; these methods struggle to impose stable constraints at material boundaries and in weak-texture regions, resulting in performance degradation in challenging self-driving scenes with high ego-motion and complex lighting.

\begin{figure*}[!t]
    \centering
    \includegraphics[width=1.0\textwidth]{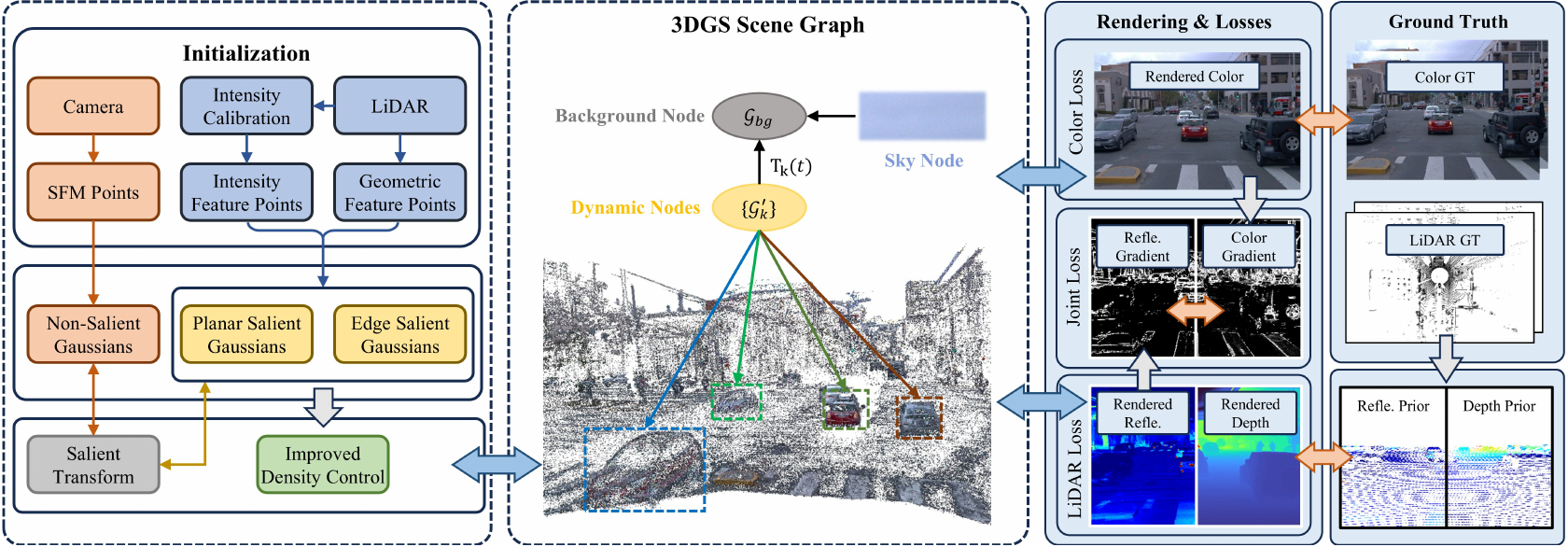}
    \caption{\textbf{Method Overview.} The initial scene Gaussians comprise Salient Gaussians from LiDAR feature points and Non-Salient Gaussians from SfM points. The scene is represented as a 3DGS scene graph with background, dynamic objects, and sky nodes. After obtaining the rendered Color, Depth, and Reflectance (Refle.) images, we optimize the scene parameters by minimizing a weighted sum of the Color, LiDAR, and Joint losses.}
    \label{Method_Overview}
\vspace{-2pt}
\end{figure*}

To address these issues, we propose a LiDAR-reflectance-guided Salient Gaussian Splatting method (LR-SGS) designed for robust reconstruction in challenging self-driving scenes, as shown in Fig. \ref{Method_Overview}. Specifically, we calibrate LiDAR intensity to ``\emph{reflectance}'' and attach it as an additional attribute channel to each Gaussian primitive, providing a new lighting-invariant material channel. Moreover, we introduce a structure-aware Salient Gaussian representation—each Salient Gaussian is endowed with a dominant direction, while the remaining two axes share a single scale, thereby reducing parameters without compromising the ability to accurately characterize contours and planar structures. To align Salient Gaussians with key structures from the outset, we initialize them from a set of LiDAR feature points comprising geometric edge points, geometric planar points, and reflectance edge points. We further develop an improved density control strategy for Salient Gaussians and a salient transform method that adaptively switches Gaussians between salient and non-salient states based on linearity and planarity of Gaussian ellipsoids. In addition, we enhance boundary consistency by aligning the gradient direction and magnitude between the reflectance and grayscale-converted RGB image, thereby reducing blur and improving reconstruction quality. Our method is validated on the Waymo Open Dataset across three representative challenging scene types, including Dense Traffic, High-Speed, and Complex Lighting, and also on static scenes. The experiments show that our method achieves consistently better results while requiring fewer Gaussians and shorter training time. In addition, we demonstrate editable reconstructions of real self-driving scenes based on our method, as shown in Fig. \ref{Overview_of_LRSGS}. Reliable scene reconstruction further enables scalable training data synthesis and realistic simulation environments, which are critical for advancing self-driving. In summary,  our contributions are as follows:

\begin{itemize}

\item We introduce LR-SGS, a robust LiDAR-reflectance-guided Salient Gaussian Splatting method for self-driving scenes that jointly optimizes geometry, appearance, and reflectance within a 3DGS scene graph.

\item We propose a structure-aware Salient Gaussian representation, initialized from LiDAR feature points and equipped with an improved density control strategy and a salient transform, achieving high-fidelity reconstruction of structural features in the environment.

\item A new lighting-invariant reflectance channel is proposed as an additional Gaussian attribute and supervision component, with direction and magnitude consistency between its gradients and grayscale RGB gradients enforced to sharpen material boundaries.

\end{itemize}

\section{Related Work}
\subsection{Driving Scene Reconstruction}

Numerous approaches have been explored for self-driving scene reconstruction. Point-based \cite{ref26,ref27} and mesh-based \cite{ref28} methods recover geometry but struggle with high-fidelity appearance. With the development of neural scene representations, several methods \cite{ref7,ref13, ref14, ref35} have employed NeRF \cite{ref2} to reconstruct self-driving scenes, but they do not handle dynamics. Compositional neural representations \cite{ref34,ref16,ref17,ref18,ref22} model dynamic vehicles but suffer from high training costs and slow rendering. 

By contrast, 3DGS \cite{ref1} achieves high-quality novel view synthesis and real-time rendering. Several studies \cite{ref10, ref23,ref24} extend 3DGS to dynamic self-driving scene reconstruction by incorporating temporal cues. HUGS \cite{ref11} achieves per-object dynamic decomposition. DeformableGS \cite{ref12} uses a set of deformable 3D Gaussians to model dynamic scenes.  PVG \cite{ref9} extends 3DGS with a time dimension for efficient large-scale dynamic scene reconstruction without annotations or pretrained models. StreetGS \cite{ref5} decomposes driving scenes into a static background and vehicle-centric dynamic Gaussians. OmniRe \cite{ref6} introduces a multi-type Gaussian scene graph that unifies rigid and non-rigid representations. These methods achieve strong temporal consistency, but under complex lighting and significant ego-motion they still struggle to recover the texture and geometry of both backgrounds and dynamic objects.

\subsection{LiDAR-Enhanced Gaussian Splatting}

3DGS \cite{ref1} initializes Gaussians from SfM points. However, this approach faces challenges in self-driving scenes, particularly under sparse viewpoints, where SfM often fails to accurately and completely recover scene geometry \cite{ref8,ref11,ref15}. Several methods \cite{ref6,ref9,ref10,ref29} use LiDAR to initialize Gaussians or as depth supervision.  StreetGS \cite{ref5}  initializes the 3DGS for both background and objects using LiDAR point clouds, while leveraging SfM points to compensate for LiDAR’s limited coverage over large areas. TCLC-GS \cite{ref25} tightly couples LiDAR and cameras by building a colorized LiDAR mesh and octree features to initialize and enrich Gaussians, and uses mesh-derived dense depth for supervision. InvRGB+L \cite{ref21} jointly reconstructs geometry, lighting, and materials in large dynamic scenes from a single camera and LiDAR sequence. 

Our method exploits the potential of LiDAR data by extracting feature points from its geometry and reflectance to initialize Salient Gaussians that capture key structures, and by integrating an RGB–LiDAR cross-modal consistency constraint to achieve more accurate reconstruction.

\section{Methodology}

\subsection{LiDAR Intensity Calibration}

LiDAR obtains the spatial coordinates and reflection intensity of target points by emitting laser pulses and recording the time-of-flight along with the returned energy of the reflected signals. The intensity $I$ can be formulated as \cite{ref36}:

\begin{equation}
	\label{eq1}
I  =\eta_{all}\frac{\rho\cos\alpha}{R^{2}}
\end{equation}
where $\eta_{all}$ denotes a LiDAR-related constant. Thus, the reflectance $\rho$ of the object can be represented using the intensity corrected by distance and angle. $R$ is the easily measurable distance. $\alpha$ is the incident angle between the object surface and the laser beam \cite{ref3}:

\begin{equation}
	\label{eq2}
\cos\alpha = \frac{\mathbf{p}^{\text{T}}\cdot\mathbf{n}}{\|\mathbf{p}\|} ,   \mathbf{n}=\frac{\left(\mathbf{p}-\mathbf{p}_{1}\right) \times\left(\mathbf{p}-\mathbf{p}_{2}\right)}{\|\mathbf{p}-\mathbf{p}_{1}\| \cdot\|\mathbf{p}-\mathbf{p}_{2}\|}
\end{equation}
where $\mathbf{n}$ denotes the local normal at point $\mathbf{p}$, $\mathbf{p}_{1}$ and $\mathbf{p}_{2}$ are neighboring points of point $\mathbf{p}$. 

Using the camera extrinsic $\mathbf{T}_{CL}$ and intrinsic $\mathbf{K}$, the point cloud is projected onto the camera plane, resulting in a sparse LiDAR reflectance image $F_{gt}$.  The reflectance values are then normalized to the range [0,1].

To further capture the material variations, we introduce the reflectance gradient. Typically, edges between different materials exhibit significant reflectance gradient. For pixel $i$, its reflectance gradient is defined as:

\begin{equation}
	\label{eq3}
g_i=\sqrt{\left(\frac{I_i-I_j}{\|\mathbf{p}_i-\mathbf{p}_j\|}\right)^2+\left(\frac{I_i-I_k}{\|\mathbf{p}_i-\mathbf{p}_k\|}\right)^2}
\end{equation}
where $I_i$ denotes the reflectance of pixel $i$, $\mathbf{p}_i$ represents its corresponding 3D point, and $j,k$ are the neighboring pixels in the horizontal and vertical directions. When a direction lacks valid neighbors, its gradient is assigned zero. Consequently, the reflectance gradient image $F_{gt}^{\prime}$ is obtained.

We employ the generated reflectance image $F_{gt}$ and its gradient image $F_{gt}^{\prime}$ as extra supervision to constrain the geometric and texture consistency of Gaussians.

\subsection{Salient Gaussians}

Self-driving scenes contain rich geometric and textural structures, such as object contours, road boundaries and ground plane. Accurate modeling of both edge and planar regions is crucial for high-fidelity reconstruction. Inspired by \cite{ref31}, we propose a scene representation guided by Salient Gaussians. Specifically, Gaussians located on scene edges gradually elongate along the edge during optimization, and their max-scale direction is regarded as the dominant direction $d_{\text{spec}}$. Gaussians in planar areas tend to flatten, with their dominant direction $d_{\text{spec}}$ corresponding to the min-scale direction. We refer to such Gaussians that characterize environmental features as Salient Gaussians. Thus, the covariance matrix $\Sigma$ of each Salient Gaussian $g$ can be formulated as: 

\begin{equation}
	\label{eq6}
\begin{cases}
\Sigma_{\text {edge }}=\mathbf{R} \operatorname{diag}\left(\sigma_{\|}^{2}, \sigma_{\perp}^{2}, \sigma_{\perp}^{2}\right) \mathbf{R}^{\text{T}} \\
\Sigma_{\text {plane }}=\mathbf{R} \operatorname{diag}\left(\sigma_{\perp}^{2}, \sigma_{\perp}^{2}, \sigma_{\|}^{2}\right) \mathbf{R}^{\text{T}} & 
\end{cases}
\end{equation}
where $\mathbf{R}$ denotes the rotation matrix converted from the quaternion $q$. The $\sigma_{\|}$ and $\sigma_{\perp}$ denote the scales along the salient and non-salient directions, respectively.

For each Salient Gaussian, we optimize a dominant scale $\sigma_{\|}$ and a shared non-dominant scale $\sigma_{\perp}$, which preserves high-fidelity geometric and textural structures while reducing optimization parameters and computational overhead. 

Building on this, we initialize the Salient Gaussians using LiDAR, which captures structured geometric features at object contours and key structural positions. Moreover, LiDAR intensity and its gradient reveal surface reflectance and texture, providing additional reflectance feature points. The geometric and reflectance feature points together form a high-confidence point set: the former are mainly distributed along edges and planes, providing global structural and contour constraints; the latter emphasize material differences and compensate for RGB weaknesses in low-texture regions. Unlike methods \cite{ref5,ref6} that initialize Gaussians directly from LiDAR points without feature extraction, we initialize Salient Gaussians from the high-confidence point set, establishing a stable scaffold at the start of training to accelerate convergence and improve reconstruction fidelity.

Specifically, we first extract geometric feature points by computing the smoothness of each LiDAR point \cite{ref32}:

\begin{equation}
	\label{eq7}
c_{j}=\frac{1}{|K| \cdot\left\|\mathbf{p}_{j}\right\|}\left\|\sum_{\mathbf{p}_{i} \in \mathcal{P}_{j}, i \neq j}\left(\mathbf{p}_{i}-\mathbf{p}_{j}\right)\right\|
\end{equation}
where $\mathcal{P}_{j}$ denotes the set of neighboring points of LiDAR point $\mathbf{p}_{j}$, containing $K$ neighbors. Based on the smoothness $c_j$, points are divided into edge and planar points to represent the edge and planar structures of the scene.

In addition to geometric features, we leverage LiDAR reflectance to extract additional edge points. For point $\mathbf{p}_{i}$, we take its left and right neighboring point sets ${\mathbf{P}}_M$ and ${\mathbf{P}}_N$ along the same ring, and compute the reflectance gradient as:

\begin{equation}
	\label{eq8}
G_j=\left|\frac{1}{M+1}\left(I_j+\sum_{{\mathbf{p}}_m^M\in{\mathbf{P}}_M}I_m\right)-\frac{1}{N}\sum_{{\mathbf{p}}_n^N\in{\mathbf{P}}_N}I_n\right|
\end{equation}
where $M$ and $N$ are the numbers of points in ${\mathbf{P}}_M$ and ${\mathbf{P}}_N$, respectively. Based on the reflectance gradient $G_{j}$, we select reflectance edge points.

We denote Salient Gaussians instantiated from geometric and reflectance edge points as Edge Salient Gaussians, and those from geometric planar points as Planar Salient Gaussians. Because LiDAR does not cover the entire scene, we initialize SfM points as Non-Salient Gaussians.

\begin{figure}[t]
	\centering
	\includegraphics[width=3.4in]{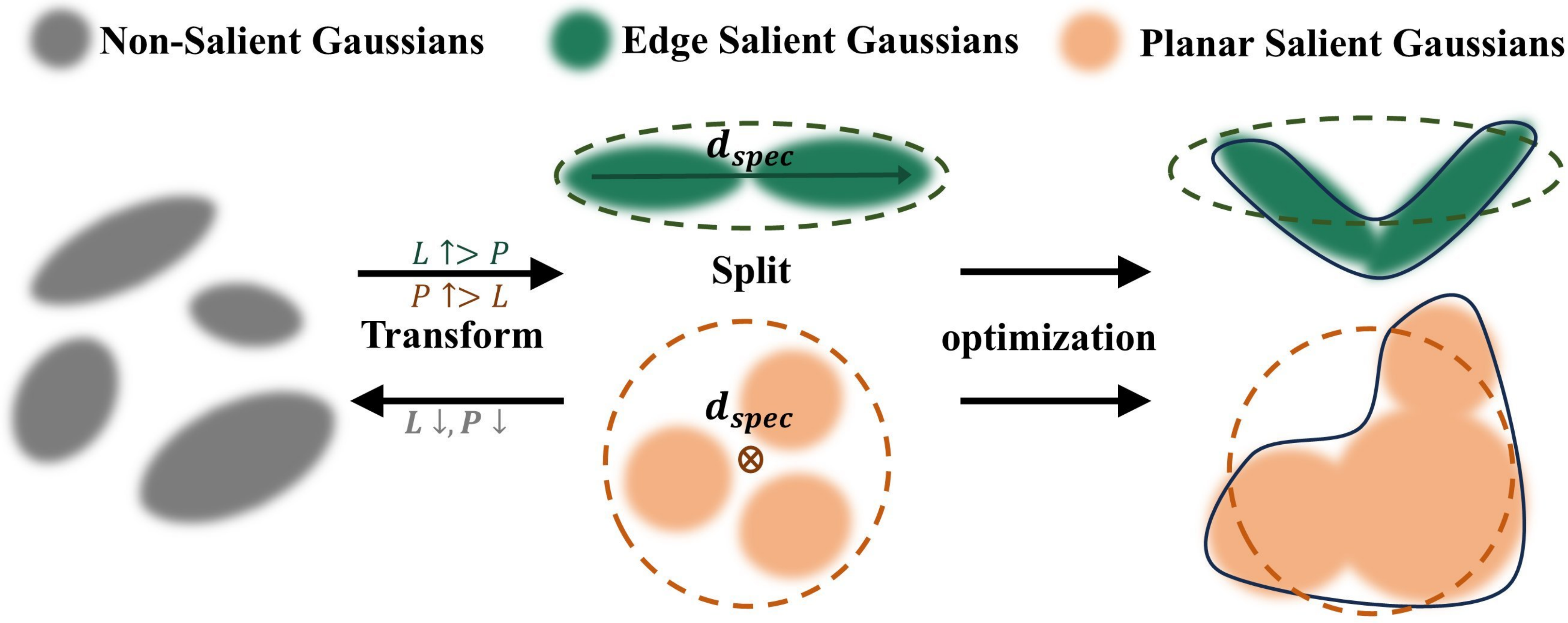}
	\caption{\textbf{Our Transform and Split.} The dashed line represents the Gaussian shape before executing split. $\uparrow$ and $\downarrow$ denote that the high and low threshold conditions are satisfied, respectively.}
	\label{Our_Transform_and_Split}
\vspace{-2pt}
\end{figure}

Moreover, we improve density control to adapt it to Salient Gaussians, as shown in Fig. \ref{Our_Transform_and_Split}. During split, Edge Salient Gaussians are split along their dominant direction, while Planar Salient Gaussians are split within the plane orthogonal to the dominant direction. New Gaussians created by split or clone inherit the type as the parent. For Non-Salient Gaussians, we follow the density control from \cite{ref6}.

To ensure adequate coverage of environmental features by Salient Gaussians, we introduce a salient transform strategy that enables mutual conversion between Salient and Non-Salient Gaussians. Specifically, Salient Gaussians that lack clear directionality are degraded, while Non-Salient Gaussians that stably capture structural features are upgraded, thereby strengthening the representation of key structures.

For a Gaussian with ordered scales $s_{1}\geq s_{2}\geq s_{3}$, we define linearity $L(g)=({s_{1}-s_{2}})/{s_{1}}$ and planarity $P(g)=({s_{2}-s_{3}})/{s_{1}}$. At each density control step, all Gaussians are evaluated. When $\max \{L(g), P(g)\}$ exceeds a set threshold $\tau_{\text{max}}$ in two successive evaluations, the Non-Salient Gaussian is upgraded to a Salient Gaussian. The type is determined by the dominant direction. When $L(g)\geq P(g)$, it becomes an Edge Salient Gaussian with scales initialized as $(\sigma_{\|}=s_{1}, \sigma_{\perp}=(s_{2}+s_{3})/2)$. Otherwise, it becomes a Planar Salient Gaussian with $(\sigma_{\perp}=(s_{1}+s_{2})/2, \sigma_{\|}=s_{3})$.

Conversely, when the salient and non-salient scales of a Salient Gaussian become similar, its ability to characterize environmental features diminishes. When ${\text{max}}\{{L(g), P(g)}\}$ remains below the set threshold $\tau_{\text{min}}$ for two consecutive evaluations, the Salient Gaussian is degraded to a non-salient one. This keeps Salient Gaussians focused on key structural regions, thereby improving reconstruction accuracy and efficiency.

\subsection{Forward Rendering}

During rendering, the color $C$, depth $D$, and reflectance $F$ of the pixel $p$ are computed by $\alpha$-blended volume rendering:

\begin{equation}
	\label{eq9}
\begin{cases}
C_\mathcal{G}=\sum_{i \in \mathcal{I}(p)} {c}_{i} \alpha_{i} \prod_{j=1}^{i-1}\left(1-\alpha_{j}\right) \\
D_\mathcal{G}=\sum_{i \in \mathcal{I}(p)} {d}_{i} \alpha_{i} \prod_{j=1}^{i-1}\left(1-\alpha_{j}\right) \\
F_\mathcal{G}=\sum_{i \in \mathcal{I}(p)} {f}_{i} \alpha_{i} \prod_{j=1}^{i-1}\left(1-\alpha_{j}\right) & 
\end{cases}
\end{equation}
where $\mathcal{I}(p)$ denotes the ordered set of Gaussians projected onto the pixel, sorted by depth from the camera center. $\alpha_{i}=o_{i} \exp \left(-\frac{1}{2}(p-\mu_{i}^{2D})^{T} \Sigma_{i}^{-1}(p-\mu_{i}^{2D})\right)$ represents the opacity weight, where $\mu_{i}^{2D}$ and $\Sigma_{i}$ are the 2D coordinates and covariance matrix, respectively.

The final rendering color is obtained by compositing the $C_\mathcal{G}$ with the $C_{sky}$ corresponding to the sky node:

\begin{equation}
	\label{eq10}
C_{\mathrm{final}}=C_{\mathcal{G}}+\left(1-O_{\mathcal{G}}\right) C_{\mathrm{sky}}
\end{equation}
where $O_{\mathcal{G}}=\sum_{i \in \mathcal{I}(p)} \alpha_{i} \prod_{j=1}^{i-1}(1-\alpha_{j})$ denotes the opacity mask of Gaussians. Since LiDAR cannot measure the sky region, the final rendered reflectance and depth are $F_{\mathrm{final}} = F_\mathcal{G}$ and $D_{\mathrm{final}} = D_\mathcal{G}$, respectively.

\subsection{Scene Optimization}

Based on the combined input of RGB and LiDAR data, we perform joint inference and optimization of the scene, enabling the simultaneous reconstruction of the geometry, appearance, and reflectance properties. Accordingly, in a single stage, we update the full set of Gaussian primitive attributes, including position, opacity, scale, rotation, appearance, and reflectance, and the parameter weights of the sky model. Finally, the scene graph $S$ is optimized using the overall loss function defined as follow:

\begin{equation}
	\label{eq11}
\mathcal{L}= \mathcal{L}_{rgb} + \mathcal{L}_{lidar} + \mathcal{L}_{joint}
\end{equation}
where $\mathcal{L}_{rgb}$ constrains photometric consistency between the rendered output and the ground truth images, $\mathcal{L}_{lidar}$ incorporates geometric and reflectance constraints from LiDAR, and $\mathcal{L}_{joint}$ ensures cross-modal consistency.

\begin{table*}[t]
  \centering
  \caption{Quantitative Comparisons on the Waymo Open Dataset.}
  \label{Quantitative_comparisons}
  \resizebox{\textwidth}{!}{%
  \begin{tabular}{lccccccccccccc}
    \toprule
    & \multicolumn{4}{c}{Dense Traffic} & \multicolumn{3}{c}{High-Speed} & \multicolumn{3}{c}{Complex Lighting} & \multicolumn{3}{c}{Static} \\
    Method & PSNR $\uparrow$ & PSNR* $\uparrow$ & SSIM $\uparrow$ & LPIPS $\downarrow$
           & PSNR $\uparrow$ & SSIM $\uparrow$ & LPIPS $\downarrow$
           & PSNR $\uparrow$ & SSIM $\uparrow$ & LPIPS $\downarrow$
           & PSNR $\uparrow$ & SSIM $\uparrow$ & LPIPS $\downarrow$\\ 
    \midrule
    3DGS* \cite{ref1}        &24.49&16.71&0.781&0.123&25.95&0.824&0.147&28.85&0.703&0.292&{28.19}&\underline{0.872}&\underline{0.121}\\
    DeformGS* \cite{ref12}   &26.08&19.55&0.817&0.102&26.02&0.839&0.139&29.01&0.717&0.285&28.14&0.863&0.124\\
    PVG \cite{ref9}          &27.95&21.67&0.841&0.093&26.24&0.845&0.139&29.12&0.709&0.284&28.05&0.854&0.126\\
    StreetGS \cite{ref5}     &27.01&21.73&0.831&0.095&28.06&\underline{0.878}&0.137&29.16&0.721&0.282&28.15&0.858&0.126\\
    OmniRe \cite{ref6}       &\underline{28.44}&\underline{23.72}&\underline{0.847}&\underline{0.085}&\underline{28.12}&0.871&\underline{0.135}&\underline{29.33}&\underline{0.727}&\underline{0.278}&\underline{28.23}&0.865&0.123\\
    \midrule
    Ours                     &\textbf{28.89}&\textbf{24.02}&\textbf{0.869}&\textbf{0.081}&\textbf{28.77}&\textbf{0.896}&\textbf{0.122}&\textbf{30.51}&\textbf{0.755}&\textbf{0.236}&\textbf{28.73}&\textbf{0.880}&\textbf{0.116}\\
    \bottomrule
  \end{tabular}}
\vspace{-6pt}
\end{table*}

\subsubsection{Color Loss}
To evaluate the photometric difference between the rendered image $C$ and the ground truth $C_{\mathrm{gt}}$, we employ a weighted combination of L1 loss and D-SSIM:

\begin{equation}
	\label{eq12}
\mathcal{L}_{rgb}=(1-\lambda_{c}) \mathcal{L}_{1}+\lambda_{c} \mathcal{L}_{D-SSIM}
\end{equation}

\subsubsection{LiDAR Loss}
LiDAR provides reliable depth and reflectance for the scene. During optimization, we utilize both its precise depth constraints and integrate reflectance and its gradient constraints, thereby providing additional constraints in weak-texture regions. The LiDAR loss is defined as:

\begin{equation}
	\label{eq13}
\mathcal{L}_{lidar}= \lambda_{depth}\mathcal{L}_{depth} + \lambda_{fle}\mathcal{L}_{fle} + \lambda_{fle}^{\prime}\mathcal{L}_{fle}^{\prime}
\end{equation}

The reflectance distortion loss $\mathcal{L}_{fle}$ is the L1 loss between the rendered reflectance $F$ and the LiDAR-projected reflectance $F_{gt}$, which enforces global reflectance consistency. In addition, we employ a mask to account for the sparsity in LiDAR observations, ensuring that constraints are applied only to valid pixels. The reflectance gradient consistency loss $\mathcal{L}_{fle}^{\prime}$ is the L1 loss between the rendered reflectance gradient obtained from Eq. \eqref{eq3} and $F_{gt}^{\prime}$. The depth distortion loss $\mathcal{L}_{depth}$ is computed in the same manner as $\mathcal{L}_{fle}$. 

\subsubsection{Joint Loss}  
To improve cross-modal consistency, we design a Joint Loss. Regions such as object boundaries, plane intersections, or material changes typically exhibit significant texture variations, making them the stable anchors for cross-modal alignment. By jointly enforcing gradient magnitude and direction consistency between LiDAR reflectance and RGB information in these regions, the reconstruction accuracy of texture boundaries and local structures is improved.

Before gradient computation, the rendered RGB image $C$ is converted to a grayscale image $C^{g}$ to eliminate interference caused by channel differences. Then, $C^{g}$ and the rendered reflectance image $F$ are Gaussian-smoothed to reduce noise while retaining the primary boundary structures:

\begin{equation}
	\label{eq14}
\tilde{I}=G_{\sigma} * I,I=(F,C^{g})
\end{equation}
where $G_{\sigma}$ is a gaussian kernel with a standard deviation of $\sigma = 1.2$ pixels. 

The gradients of the smoothed images $\tilde{I}$ are computed using the Scharr kernel, with mirrored padding at the edges, yielding the following gradient:

\begin{equation}
	\label{eq15}
g_{I}=\sqrt{I_{x}^{2}+I_{y}^{2}}
\end{equation}
where $I_{x}$ and $I_{y}$ represent the gradient components in the horizontal and vertical directions, respectively. 

The Joint Loss consists of two components: direction consistency and magnitude consistency, as follows:

\begin{equation}
	\label{eq16}
\mathcal{L}_{joint}= \lambda_{dir}\mathcal{L}_{dir} + \lambda_{val}\mathcal{L}_{val}
\end{equation}
where $\mathcal{L}_{dir}=1-(\hat{\nabla} F \cdot \hat{\nabla} C^{g})$ is the gradient direction consistency loss, where $\hat{\nabla} F = \left(F_{x}, F_{y}\right)/g_{F}$ and $\hat{\nabla} C^{g} = \left(C_{x}^{g}, C_{y}^{g}\right)/g_{C^{g}}$ are the normalized gradient vectors of the reflectance and grayscale images, respectively, used to enforce consistency in edge orientation. $\mathcal{L}_{val}=\left\| g_{F}/{F} - g_{C^{g}}/{C^{g}} \right\|_{1}$ denotes the normalized magnitude consistency loss, which removes cross-modal scale differences via normalization while preserving edge prominence.

\section{Experiments}
\subsection{Experimental Setup}
\subsubsection{Datasets}

We conduct experiments on the Waymo Open Dataset \cite{ref30}, containing RGB images from five cameras and 64-beam LiDAR data with intensity. To thoroughly assess our method, we select 24 sequences from the Waymo Open Dataset across four categories: Dense Traffic, High-Speed, Complex Lighting, and Static, with six sequences for each category. Every fourth frame is used for testing, and the rest for training.

\subsubsection{Baselines}

We compare our method against several state-of-the-art methods, including 3DGS \cite{ref1}, DeformGS \cite{ref12}, PVG \cite{ref9}, StreetGS \cite{ref5}, and OmniRe \cite{ref6}. We evaluate RGB rendering using PSNR, SSIM, and LPIPS \cite{ref33}, while for LiDAR reflectance, we assess performance using RMSE. We report PSNR on moving objects (PSNR*) in Dense Traffic scenes to assess reconstruction of dynamic objects.

\subsubsection{Implementation Details}

All experiments are conducted for 30k training iterations. We use the implementations of 3DGS and DeformGS with LiDAR depth supervision (3DGS* and DeformGS*). Object masks are obtained following \cite{ref21}. The salient transform thresholds are set as hyperparameters to $\tau_{\text{max}}=0.5$ and $\tau_{\text{min}}=0.1$. Loss weights are set to $\lambda_{c}=\lambda_{val}=0.2$, $\lambda_{depth}=\lambda_{fle}=\lambda_{dir}=0.1$, and $\lambda_{fle}^{\prime}=0.05$.

\begin{figure*}[!t]
    \centering
    \includegraphics[width=1.0\textwidth]{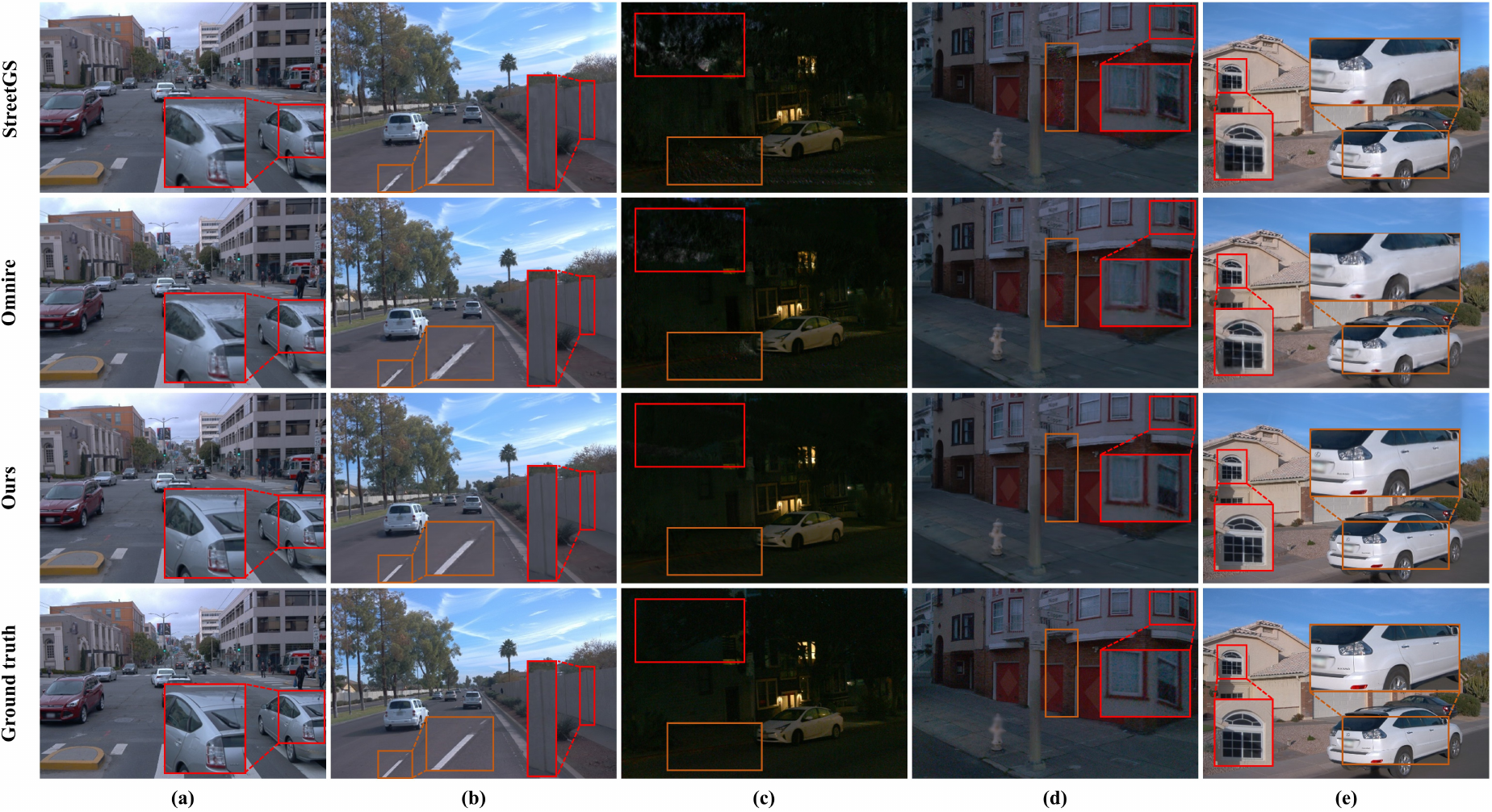}
    \caption{\textbf{Qualitative Comparison of Novel View Synthesis.} (a) shows the Dense Traffic scene. (b) shows the High-Speed scene. (c) and (d) show the scene with Complex Lighting conditions. (e) shows the Static scene. Our method not only achieves high-quality reconstruction of dynamic objects, but also recovers very fine details of the background environment, maintaining consistent and stable reconstructions even under complex lighting conditions.}
    \label{Qualitative_Comparison_of_Novel_View_Synthesis}
\vspace{-2pt}
\end{figure*}

\subsection{Qualitative and Quantitative Results}

Table \ref{Quantitative_comparisons} presents quantitative results for novel view synthesis comparing our method with the baselines. The best and second-best results are highlighted in bold and underlined, respectively. Across all metrics, our method achieves competitive performance over the different scene categories, demonstrating stronger adaptability and generalization.

Fig. \ref{Qualitative_Comparison_of_Novel_View_Synthesis} shows qualitative results on the Waymo Open Dataset. In scenes with dynamic objects, benefiting from the cooperation of the Salient Gaussians and LiDAR reflectance constraints, our method recovers object appearance more faithfully. As shown in Fig. \ref{Qualitative_Comparison_of_Novel_View_Synthesis}(a), the taillights and roof outline of the gray vehicle are clearly reconstructed, while StreetGS and OmniRe exhibit blurred artifacts. Additionally, under high-speed that lowers co-visibility across frames, our approach is more sensitive to geometric and textural boundaries, as shown in Fig. \ref{Qualitative_Comparison_of_Novel_View_Synthesis}(b), capturing details such as wall protrusions and lane markings. Under complex lighting conditions, such as the nighttime scene in Fig. \ref{Qualitative_Comparison_of_Novel_View_Synthesis}(c), lighting-invariant LiDAR reflectance provides stable constraints, effectively mitigating artifacts in StreetGS and OmniRe and maintaining consistent scene reconstruction. For the static scene in Fig. \ref{Qualitative_Comparison_of_Novel_View_Synthesis}(e), Salient Gaussians establish stronger priors on critical geometric and textural structures, and together with complementary texture information and boundary cues from reflectance, improve global consistency and detail recovery in static environment. Overall, our method achieves higher-fidelity reconstruction of dynamic objects and static environments, producing novel view synthesis rich in geometric and textural details even under challenging conditions.

\begin{figure}[t]
	\centering
	\includegraphics[width=3.4in]{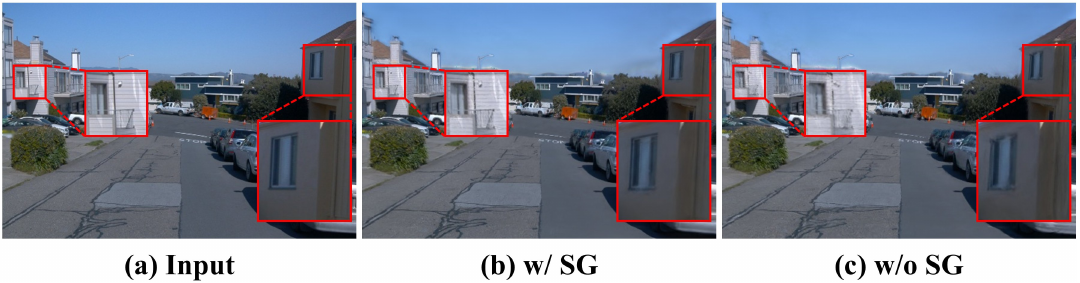}
	\caption{\textbf{Ablation study on Salient Gaussians.} Salient Gaussians enable finer reconstruction of structural features in the environment.}
	\label{Ablation_study_on_Salient_Gaussian}
\vspace{-2pt}
\end{figure}

\begin{table}[t]
  \centering
  \caption{Ablation study of each component on the Waymo Open Dataset.}
  \label{Ablation_study}
  \begin{tabular*}{\linewidth}{@{\extracolsep{\fill}}lccc}
    \toprule
    Method              & PSNR $\uparrow$ & SSIM $\uparrow$ & LPIPS $\downarrow$ \\
    \midrule
    w/o SG              & 28.74           & 0.830           & 0.152 \\
    w/o LF-Init         & 28.94           & 0.839           & 0.144 \\
    w/o Reflectance     & 28.87           & 0.831           & 0.147 \\
    w/o Joint Loss      & 28.96           & 0.835           & 0.144 \\
    Ours                & \textbf{29.22}  & \textbf{0.850}  & \textbf{0.139} \\
    \bottomrule
  \end{tabular*}
\vspace{-6pt}
\end{table}

\begin{table}[t]
  \centering
  \caption{Ablation study on the static scene.}
  \label{Ablation_study_on_the_static_Scene}
  \resizebox{\linewidth}{!}{
    \begin{tabular}{c|l|llll}
      \toprule
      \multicolumn{1}{l|}{Iteration} & \multicolumn{1}{c|}{Method} & PSNR $\uparrow$ & SSIM $\uparrow$ & LPIPS $\downarrow$ & Training $\downarrow$ \\
      \midrule
      \multirow{3}{*}{7k}  & w/o LF-Init  & 25.50 & 0.816 & 0.213 & 13m12s \\
                           & w/o SG    & 25.52 & 0.824 & 0.209 & 13m48s \\
                           & Ours      & \textbf{26.41} & \textbf{0.839} & \textbf{0.188} & \textbf{11m05s} \\
      \midrule
      \multirow{3}{*}{15k} & w/o LF-Init  & 27.98 & 0.860 & 0.140 & 30m34s \\
                           & w/o SG    & 27.54 & 0.858 & 0.143 & 31m17s \\
                           & Ours      & \textbf{28.13} & \textbf{0.863} & \textbf{0.136} & \textbf{28m54s} \\
      \midrule
      \multirow{3}{*}{30k} & w/o LF-Init  & 28.76 & 0.890 & 0.115 & 68m37s \\
                           & w/o SG    & 28.51 & 0.874 & 0.119 & 70m21s \\
                           & Ours      & \textbf{28.96} & \textbf{0.894} & \textbf{0.112} & \textbf{61m57s} \\
      \bottomrule
    \end{tabular}
  }
\vspace{-6pt}
\end{table}

\subsection{Ablation Study}

We validate the design choices of our method on all selected sequences of the Waymo Open Dataset. Table \ref{Ablation_study} presents the quantitative results averaged over all sequences.

\subsubsection{Salient Gaussian}

We conduct an ablation by removing Salient Gaussians and using LiDAR feature points to initialize Non-Salient Gaussians, in order to assess the effectiveness of Salient Gaussians, denoted as w/o SG. Table \ref{Ablation_study_on_the_static_Scene} reports novel view synthesis metrics and training time. The results show that introducing Salient Gaussians improves rendering quality, accelerates convergence, and reduces training time. This improvement arises because Salient Gaussians better match edges and planar structures in the environment and require fewer parameters. Fig. \ref{Ablation_study_on_Salient_Gaussian}(b) and \ref{Ablation_study_on_Salient_Gaussian}(c) compare reconstructions with and without Salient Gaussians. Even in regions not covered by LiDAR, structural details are well reconstructed. This benefit stems from our salient transform strategy, which enables Salient Gaussians to grow beyond their initial coverage and distribute across entire scene.

\begin{figure}[t]
	\centering
	\includegraphics[width=3.4in]{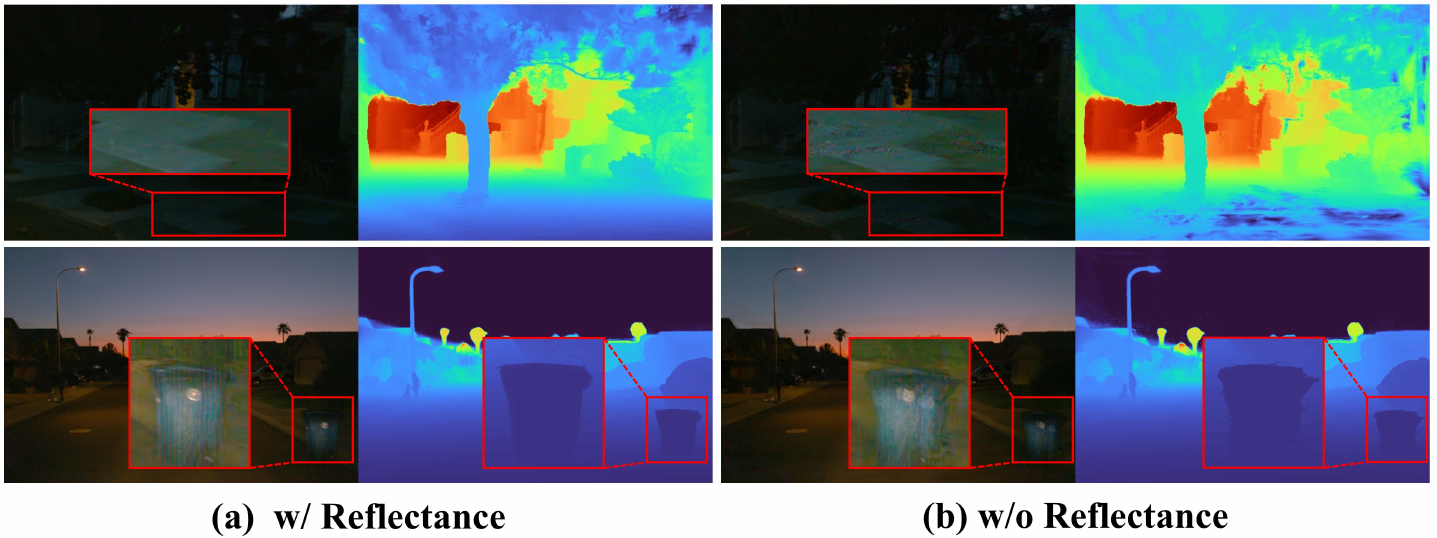}
	\caption{\textbf{Ablation study on LiDAR Reflectance.} We show the rendered image and depth of our method with and without LiDAR Reflectance. For clearer visual comparison, we increased the brightness in the zoomed-in regions.}

	\label{Ablation_of_LiDAR_Reflectance}
\vspace{-2pt}
\end{figure}

\begin{figure}[t]
	\centering
	\includegraphics[width=3.4in]{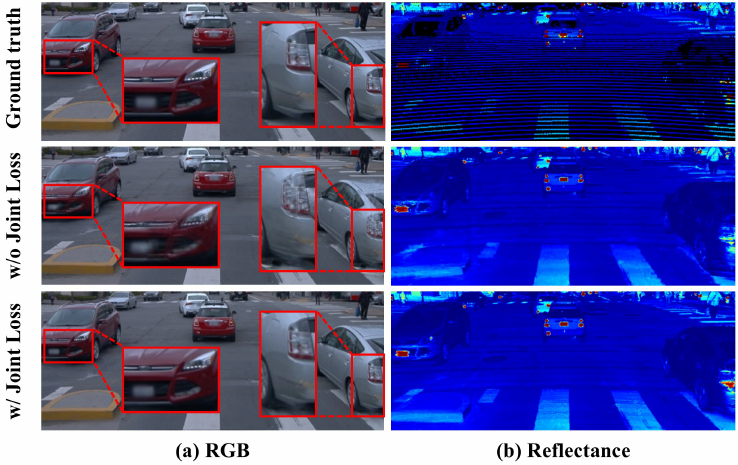}
	\caption{\textbf{Ablation of the Joint Loss.} (a) the rendered image. (b) the rendered reflectance.}

	\label{Ablation_of_Joint_Loss}
\vspace{-2pt}
\end{figure}

\begin{table}[t]
  \centering
  \caption{Ablation Study on Joint Loss.}
  \label{Ablation_Study_on_Joint_Loss}
  {\setlength{\tabcolsep}{10pt}
  \begin{tabular*}{\linewidth}{@{\extracolsep{\fill}}lcccc}
    \toprule
    Method           & PSNR $\uparrow$ & SSIM $\uparrow$ & LPIPS $\downarrow$ & RMSE $\downarrow$ \\
    \midrule
    w/o Joint Loss   & 30.08           & 0.913           & 0.057              & 0.1063 \\
    w/ Joint Loss    & \textbf{30.39}  & \textbf{0.917}  & \textbf{0.053}     & \textbf{0.0854} \\
    \bottomrule
  \end{tabular*}}
\vspace{-6pt}
\end{table}

\begin{table}[t]
  \centering
  \caption{Efficiency Analysis.}
  \label{Efficiency_analysis}
  {\setlength{\tabcolsep}{10pt}
  \begin{tabular*}{\linewidth}{@{\extracolsep{\fill}}lcccc}
    \toprule
    Method & PSNR $\uparrow$ & Number $\downarrow$ & Training $\downarrow$ & FPS $\uparrow$ \\
    \midrule
    StreetGS \cite{ref5} & 28.20 				& 2,929,851 					& \underline{64m30s} & \underline{33.61}\\ 
    OmniRe \cite{ref6}   & \underline{28.62} & \underline{2,744,275} 	& 67m11s 					&30.55\\				
    Ours                 & \textbf{28.95} 	& \textbf{2,510,883} 		& \textbf{59m25s} 		&\textbf{36.95}\\	
    \bottomrule
  \end{tabular*}}
\vspace{-6pt}
\end{table}

\subsubsection{LiDAR Feature Initialization}

We evaluate the impact of LiDAR feature-point initialization (LF-Init) of Salient Gaussians on reconstruction. Without LF-Init, we bypass feature extraction, initialize Non-Salient Gaussians from raw LiDAR and SfM points, and rely solely on the salient transform strategy to produce Salient Gaussians during training. Table \ref{Ablation_study} shows that initializing Salient Gaussians with LiDAR feature points improves rendered image quality. Moreover, Table \ref{Ablation_study_on_the_static_Scene} shows that such initialization provides a better starting point for Salient Gaussians to align with critical geometric and textural structures early on, yielding faster convergence and higher reconstruction accuracy.

\subsubsection{LiDAR Reflectance}

We evaluate the impact of LiDAR reflectance by comparing our full method to a variant using only Color Loss and LiDAR Depth Loss. Table \ref{Ablation_study} shows that introducing LiDAR reflectance as an additional Gaussian attribute, together with the associated supervision, significantly improves rendering quality. Moreover, Fig. \ref{Ablation_of_LiDAR_Reflectance} indicates that the lighting-invariant LiDAR reflectance supplies stable constraints under complex lighting conditions, enabling more accurate geometric reconstruction and fewer blurry artifacts.

\subsubsection{Joint Loss}

To evaluate the effect of the proposed Joint Loss, we conduct an ablation study where only the RGB Loss and LiDAR Loss terms are used. Specifically, we disable the Joint Loss to decouple the mutual influence between RGB and LiDAR during optimization. Table \ref{Ablation_Study_on_Joint_Loss} provides ablation studies of the Joint Loss  on the dynamic scene. As shown in Fig. \ref{Ablation_of_Joint_Loss}, vehicles appear blurry without Joint Loss, whereas introducing Joint Loss preserves clear vehicle contours, indicating that the combined textural cues from RGB and LiDAR reflectance effectively improve overall reconstruction accuracy. Furthermore, the Joint Loss makes the rendered reflectance closer to the Ground Truth. For example, highlights in the license-plate and front-light regions are reconstructed more accurately, and crosswalk and platform boundaries become clearer and more consistently aligned.

\subsection{Efficiency Analysis}

We select four sequences, one from each scene category, to evaluate the number of Gaussians, training time, and FPS. Table \ref{Efficiency_analysis} reports the mean results across these four sequences. Compared with the baseline methods, our method achieves the best performance. This advantage is attributed to the well-initialized Salient Gaussians that capture critical scene structures, thereby suppressing redundant Gaussians. Moreover, the reduced parameterization of Salient Gaussians further shortens training time. Our approach achieves improved reconstruction quality while also improving efficiency.

\section{Conclusion}

In this work, we proposed a robust LiDAR-reflectance-guided Salient Gaussian Splatting method (LR-SGS) tailored for self-driving scenes. By calibrating LiDAR intensity to obtain a lighting-invariant reflectance channel, initializing structure-aware Salient Gaussians from LiDAR geometric and reflectance feature points, LR-SGS achieves improved fidelity and robustness under challenging conditions. Moreover, the editable scene representations of LR-SGS enable scalable training data generation and realistic simulation environments for self-driving. We believe our work significantly contributes to advancing self-driving. Future work will cover broader scenarios and larger-scale self-driving scenes.


\bibliographystyle{IEEEtran} 

\bibliography{ref}

\begin{thebibliography}{10}
\providecommand{\url}[1]{#1}
\csname url@samestyle\endcsname
\providecommand{\newblock}{\relax}
\providecommand{\bibinfo}[2]{#2}
\providecommand{\BIBentrySTDinterwordspacing}{\spaceskip=0pt\relax}
\providecommand{\BIBentryALTinterwordstretchfactor}{4}
\providecommand{\BIBentryALTinterwordspacing}{\spaceskip=\fontdimen2\font plus
\BIBentryALTinterwordstretchfactor\fontdimen3\font minus
  \fontdimen4\font\relax}
\providecommand{\BIBforeignlanguage}[2]{{%
\expandafter\ifx\csname l@#1\endcsname\relax
\typeout{** WARNING: IEEEtran.bst: No hyphenation pattern has been}%
\typeout{** loaded for the language `#1'. Using the pattern for}%
\typeout{** the default language instead.}%
\else
\language=\csname l@#1\endcsname
\fi
#2}}
\providecommand{\BIBdecl}{\relax}
\BIBdecl

\bibitem{ref34}
\BIBentryALTinterwordspacing
J.~Yang, B.~Ivanovic, O.~Litany, X.~Weng, S.~W. Kim, B.~Li, T.~Che, D.~Xu,
  S.~Fidler, M.~Pavone, and Y.~Wang, ``Emerne{RF}: Emergent spatial-temporal
  scene decomposition via self-supervision,'' in \emph{The Twelfth
  International Conference on Learning Representations}, 2024. [Online].
  Available: \url{https://openreview.net/forum?id=ycv2z8TYur}
\BIBentrySTDinterwordspacing

\bibitem{ref1}
\BIBentryALTinterwordspacing
B.~Kerbl, G.~Kopanas, T.~Leimk{\"u}hler, and G.~Drettakis, ``3d gaussian
  splatting for real-time radiance field rendering,'' \emph{ACM Transactions on
  Graphics}, vol.~42, no.~4, July 2023. [Online]. Available:
  \url{https://repo-sam.inria.fr/fungraph/3d-gaussian-splatting/}
\BIBentrySTDinterwordspacing

\bibitem{ref8}
S.~Yu, C.~Cheng, Y.~Zhou, X.~Yang, and H.~Wang, ``Rgb-only gaussian splatting
  slam for unbounded outdoor scenes,'' in \emph{2025 IEEE International
  Conference on Robotics and Automation (ICRA)}, 2025, pp. 11\,068--11\,074.

\bibitem{ref11}
H.~Zhou, J.~Shao, L.~Xu, D.~Bai, W.~Qiu, B.~Liu, Y.~Wang, A.~Geiger, and
  Y.~Liao, ``Hugs: Holistic urban 3d scene understanding via gaussian
  splatting,'' in \emph{Proceedings of the IEEE/CVF Conference on Computer
  Vision and Pattern Recognition (CVPR)}, June 2024, pp. 21\,336--21\,345.

\bibitem{ref15}
Z.~Xin, C.~Wu, P.~Huang, Y.~Zhang, Y.~Mao, and G.~Huang, ``Large-scale gaussian
  splatting slam,'' in \emph{2025 IEEE International Conference on Robotics and
  Automation (ICRA)}, 2025, pp. 8478--8485.

\bibitem{ref37}
F.~Zhu, Y.~Zhao, Z.~Chen, B.~Yu, and H.~Zhu, ``Fgo-slam: Enhancing gaussian
  slam with globally consistent opacity radiance field,'' in \emph{2025 IEEE
  International Conference on Robotics and Automation (ICRA)}.\hskip 1em plus
  0.5em minus 0.4em\relax IEEE, 2025, pp. 11\,075--11\,081.

\bibitem{ref3}
H.~Wang, C.~Wang, and L.~Xie, ``Intensity-slam: Intensity assisted localization
  and mapping for large scale environment,'' \emph{IEEE Robotics and Automation
  Letters}, vol.~6, no.~2, pp. 1715--1721, 2021.

\bibitem{ref4}
Y.~Zhang, Y.~Tian, W.~Wang, G.~Yang, Z.~Li, F.~Jing, and M.~Tan, ``Ri-lio:
  Reflectivity image assisted tightly-coupled lidar-inertial odometry,''
  \emph{IEEE Robotics and Automation Letters}, vol.~8, no.~3, pp. 1802--1809,
  2023.

\bibitem{ref5}
Y.~Yan, H.~Lin, C.~Zhou, W.~Wang, H.~Sun, K.~Zhan, X.~Lang, X.~Zhou, and
  S.~Peng, ``Street gaussians: Modeling dynamic urban scenes with gaussian
  splatting,'' in \emph{ECCV}, 2024.

\bibitem{ref6}
Z.~Chen, J.~Yang, J.~Huang, R.~de~Lutio, J.~M. Esturo, B.~Ivanovic, O.~Litany,
  Z.~Gojcic, S.~Fidler, M.~Pavone, L.~Song, and Y.~Wang, ``Omnire: Omni urban
  scene reconstruction,'' in \emph{The Thirteenth International Conference on
  Learning Representations}, 2025.

\bibitem{ref9}
\BIBentryALTinterwordspacing
Y.~Chen, C.~Gu, J.~Jiang, X.~Zhu, and L.~Zhang, ``Periodic vibration gaussian:
  Dynamic urban scene reconstruction and real-time rendering,'' \emph{CoRR},
  vol. abs/2311.18561, 2023. [Online]. Available:
  \url{https://doi.org/10.48550/arXiv.2311.18561}
\BIBentrySTDinterwordspacing

\bibitem{ref10}
X.~Zhou, Z.~Lin, X.~Shan, Y.~Wang, D.~Sun, and M.-H. Yang, ``Drivinggaussian:
  Composite gaussian splatting for surrounding dynamic autonomous driving
  scenes,'' in \emph{Proceedings of the IEEE/CVF conference on computer vision
  and pattern recognition}, 2024, pp. 21\,634--21\,643.

\bibitem{ref26}
J.~Ost, I.~Laradji, A.~Newell, Y.~Bahat, and F.~Heide, ``Neural point light
  fields,'' in \emph{Proceedings of the IEEE/CVF Conference on Computer Vision
  and Pattern Recognition}, 2022, pp. 18\,419--18\,429.

\bibitem{ref27}
K.~Rematas, A.~Liu, P.~P. Srinivasan, J.~T. Barron, A.~Tagliasacchi,
  T.~Funkhouser, and V.~Ferrari, ``Urban radiance fields,'' in
  \emph{Proceedings of the IEEE/CVF Conference on Computer Vision and Pattern
  Recognition}, 2022, pp. 12\,932--12\,942.

\bibitem{ref28}
J.~Y. Liu, Y.~Chen, Z.~Yang, J.~Wang, S.~Manivasagam, and R.~Urtasun,
  ``Real-time neural rasterization for large scenes,'' in \emph{Proceedings of
  the IEEE/CVF International Conference on Computer Vision}, 2023, pp.
  8416--8427.

\bibitem{ref7}
X.~Zhong, Y.~Pan, J.~Behley, and C.~Stachniss, ``Shine-mapping: Large-scale 3d
  mapping using sparse hierarchical implicit neural representations,'' in
  \emph{2023 IEEE International Conference on Robotics and Automation
  (ICRA)}.\hskip 1em plus 0.5em minus 0.4em\relax IEEE, 2023, pp. 8371--8377.

\bibitem{ref13}
F.~Lu, Y.~Xu, G.~Chen, H.~Li, K.-Y. Lin, and C.~Jiang, ``Urban radiance field
  representation with deformable neural mesh primitives,'' in \emph{Proceedings
  of the IEEE/CVF International Conference on Computer Vision}, 2023, pp.
  465--476.

\bibitem{ref14}
X.~Zhang, A.~Kundu, T.~Funkhouser, L.~Guibas, H.~Su, and K.~Genova, ``Nerflets:
  Local radiance fields for efficient structure-aware 3d scene representation
  from 2d supervision,'' in \emph{Proceedings of the IEEE/CVF Conference on
  Computer Vision and Pattern Recognition}, 2023, pp. 8274--8284.

\bibitem{ref35}
C.~Jiang, R.~Niu, Z.~Chen, C.~Hua, X.~Kuang, X.~Fu, B.~Yu, and H.~Zhu,
  ``Large-scale neural scene disentanglement approach for self-driving
  simulation,'' \emph{IEEE Transactions on Intelligent Vehicles}, pp. 1--12,
  2024.

\bibitem{ref2}
B.~Mildenhall, P.~P. Srinivasan, M.~Tancik, J.~T. Barron, R.~Ramamoorthi, and
  R.~Ng, ``Nerf: Representing scenes as neural radiance fields for view
  synthesis,'' in \emph{ECCV}, 2020.

\bibitem{ref16}
Z.~Wu, T.~Liu, L.~Luo, Z.~Zhong, J.~Chen, H.~Xiao, C.~Hou, H.~Lou, Y.~Chen,
  R.~Yang \emph{et~al.}, ``Mars: An instance-aware, modular and realistic
  simulator for autonomous driving,'' in \emph{CAAI International Conference on
  Artificial Intelligence}.\hskip 1em plus 0.5em minus 0.4em\relax Springer,
  2023, pp. 3--15.

\bibitem{ref17}
Z.~Yang, Y.~Chen, J.~Wang, S.~Manivasagam, W.-C. Ma, A.~J. Yang, and
  R.~Urtasun, ``Unisim: A neural closed-loop sensor simulator,'' in
  \emph{Proceedings of the IEEE/CVF Conference on Computer Vision and Pattern
  Recognition}, 2023, pp. 1389--1399.

\bibitem{ref18}
H.~Turki, J.~Y. Zhang, F.~Ferroni, and D.~Ramanan, ``Suds: Scalable urban
  dynamic scenes,'' in \emph{Proceedings of the IEEE/CVF Conference on Computer
  Vision and Pattern Recognition}, 2023, pp. 12\,375--12\,385.

\bibitem{ref22}
T.~Wu, F.~Zhong, A.~Tagliasacchi, F.~Cole, and C.~Oztireli, ``D\^{} 2nerf:
  Self-supervised decoupling of dynamic and static objects from a monocular
  video,'' \emph{Advances in neural information processing systems}, vol.~35,
  pp. 32\,653--32\,666, 2022.

\bibitem{ref23}
Z.~Yang, H.~Yang, Z.~Pan, and L.~Zhang, ``Real-time photorealistic dynamic
  scene representation and rendering with 4d gaussian splatting,'' in
  \emph{ICLR}, 2024.

\bibitem{ref24}
G.~Wu, T.~Yi, J.~Fang, L.~Xie, X.~Zhang, W.~Wei, W.~Liu, Q.~Tian, and X.~Wang,
  ``4d gaussian splatting for real-time dynamic scene rendering,'' in
  \emph{Proceedings of the IEEE/CVF conference on computer vision and pattern
  recognition}, 2024, pp. 20\,310--20\,320.

\bibitem{ref12}
Z.~Yang, X.~Gao, W.~Zhou, S.~Jiao, Y.~Zhang, and X.~Jin, ``Deformable 3d
  gaussians for high-fidelity monocular dynamic scene reconstruction,'' in
  \emph{Proceedings of the IEEE/CVF conference on computer vision and pattern
  recognition}, 2024, pp. 20\,331--20\,341.

\bibitem{ref29}
J.~Shen, H.~Yu, J.~Wu, W.~Yang, and G.-S. Xia, ``Lidar-enhanced 3d gaussian
  splatting mapping,'' in \emph{2025 IEEE International Conference on Robotics
  and Automation (ICRA)}, 2025, pp. 2048--2054.

\bibitem{ref25}
C.~Zhao, S.~Sun, R.~Wang, Y.~Guo, J.-J. Wan, Z.~Huang, X.~Huang, Y.~V. Chen,
  and L.~Ren, ``Tclc-gs: Tightly coupled lidar-camera gaussian splatting for
  autonomous driving: Supplementary materials,'' in \emph{European Conference
  on Computer Vision}.\hskip 1em plus 0.5em minus 0.4em\relax Springer, 2024,
  pp. 91--106.

\bibitem{ref21}
X.~Chen, B.~Chandaka, C.-H. Lin, Y.-Q. Zhang, D.~Forsyth, H.~Zhao, and S.~Wang,
  ``Invrgb+ l: Inverse rendering of complex scenes with unified color and lidar
  reflectance modeling,'' \emph{arXiv preprint arXiv:2507.17613}, 2025.

\bibitem{ref36}
D.~P. Frost, O.~Kähler, and D.~W. Murray, ``Object-aware bundle adjustment for
  correcting monocular scale drift,'' in \emph{2016 IEEE International
  Conference on Robotics and Automation (ICRA)}, 2016, pp. 4770--4776.

\bibitem{ref31}
S.~Liu, T.~Deng, H.~Zhou, L.~Li, H.~Wang, D.~Wang, and M.~Li, ``Mg-slam:
  Structure gaussian splatting slam with manhattan world hypothesis,''
  \emph{IEEE Transactions on Automation Science and Engineering}, vol.~22, pp.
  17\,034--17\,049, 2025.

\bibitem{ref32}
\BIBentryALTinterwordspacing
J.~Zhang and S.~Singh, ``Loam: Lidar odometry and mapping in real-time,'' in
  \emph{Robotics: Science and Systems}, 2014. [Online]. Available:
  \url{https://api.semanticscholar.org/CorpusID:18612391}
\BIBentrySTDinterwordspacing

\bibitem{ref30}
P.~Sun, H.~Kretzschmar, X.~Dotiwalla, A.~Chouard, V.~Patnaik, P.~Tsui, J.~Guo,
  Y.~Zhou, Y.~Chai, B.~Caine, V.~Vasudevan, W.~Han, J.~Ngiam, H.~Zhao,
  A.~Timofeev, S.~Ettinger, M.~Krivokon, A.~Gao, A.~Joshi, Y.~Zhang, J.~Shlens,
  Z.~Chen, and D.~Anguelov, ``Scalability in perception for autonomous driving:
  Waymo open dataset,'' in \emph{2020 IEEE/CVF Conference on Computer Vision
  and Pattern Recognition (CVPR)}, 2020, pp. 2443--2451.

\bibitem{ref33}
R.~Zhang, P.~Isola, A.~A. Efros, E.~Shechtman, and O.~Wang, ``The unreasonable
  effectiveness of deep features as a perceptual metric,'' in \emph{Proceedings
  of the IEEE conference on computer vision and pattern recognition}, 2018, pp.
  586--595.

\end{thebibliography}


\end{document}